\pdfoutput=1
\documentclass[11pt]{article}

\usepackage[preprint]{acl}

\usepackage{times}
\usepackage{latexsym}
\usepackage{times}
\usepackage{latexsym}
\usepackage{xstring}
\usepackage{xcolor}
\usepackage{color}
\usepackage{ulem}
\usepackage{bbm}
\usepackage{amsmath}
\usepackage{amsfonts}
\usepackage{makecell}
\usepackage{algorithm}
\usepackage{algorithmic}
\usepackage{booktabs}
\usepackage{framed}
\usepackage{pifont}
\usepackage{graphicx}

\usepackage[T1]{fontenc}
\usepackage[utf8]{inputenc}

\usepackage{microtype}

\usepackage{inconsolata}

\title{Large, Small or Both: A Novel Data Augmentation Framework Based on Language Models for Debiasing Opinion Summarization}

\author{\bf{Yanyue Zhang}$^{\spadesuit}$, \bf{Pengfei Li}$^{\spadesuit}$, \bf{Yilong Lai}$^{\spadesuit}$, \bf{Deyu Zhou\thanks{\;Corresponding author.}}$^{\spadesuit}$ and \bf{Yulan He}$^{\heartsuit}$\\
$^{\spadesuit}$School of Computer Science and Engineering, Key Laboratory of Computer Network \\ and Information Integration, Ministry of Education, Southeast University, China \\
$^{\heartsuit}$Department of Informatics, King’s College London
$^{\heartsuit}$The Alan Turing Institute \\
\texttt{\{yanyuez98,lip.f,yilong.lai,d.zhou\}@seu.edu.cn},\\
  \texttt{yulan.he@kcl.ac.uk}}

\begin{document}
\maketitle
\begin{abstract}
As more than 70$\%$ of reviews in the existing opinion summary data set are positive, current opinion summarization approaches are reluctant to generate negative summaries given the input of negative texts. 
To address such sentiment bias, a direct approach without the over-reliance on a specific framework is to generate additional data based on large language models to balance the emotional distribution of the dataset. However, data augmentation based on large language models faces two disadvantages: 1) the potential issues or toxicity in the augmented data; 2) the expensive costs. 
Therefore, in this paper, we propose a novel data augmentation framework based on both large and small language models for debiasing opinion summarization.
In specific, a small size of synthesized negative reviews is obtained by rewriting the positive text via a large language model. Then, a disentangle reconstruction model is trained based on the generated data.
After training, a large amount of synthetic data can be obtained by decoding the new representation obtained from the combination of different sample representations and filtering based on confusion degree and sentiment classification.  
Experiments have proved that our framework can effectively alleviate emotional bias same as using only large models, but more economically.

\end{abstract}

\section{Introduction}

With the unprecedented development of online interactive platforms, reviews on shopping platforms or social media become an important information source for manufacturers to make decisions. To cope with the flood of reviews, opinion summarization has received significant interest in natural language processing communities. Unlike other summarization tasks for news, Wikipedia, and medical treatment records, opinion summarization focuses on texts with user opinions and subjective emotions about an entity (e.g., a product, hotel, or restaurant). Accurately summarizing user perceptions and attitudes towards entities is a core requirement of opinion summarization.

\begin{table}[!t]
\setlength{\belowcaptionskip}{-0.5cm}
\scalebox{0.90}{
\begin{tabular}{p{230pt}}
\hline
\textbf{Reviews:} \\
  \ding{172} The tights are \textcolor{red}{badly made} and \textcolor{red}{can't last several washings (hang dry). The color is ugly, and my daughter hates $\dots$} \ding{173} $\dots$ common ballet tights. They \textcolor{red}{can't fit well} and \textcolor{red}{squish her toes as much as some others}. $\dots$ \ding{174} my 3 year old  \textcolor{red}{can't fit into these perfectly}. $\dots$\ding{175} \textcolor{red}{Stiff fabric}, runs \textcolor{red}{small} a though. $\dots$\ding{176} \textcolor{red}{This is not my go to tight when my daughter needs new ones.} \\ 
\hline 
\textbf{Summary by Coop:} \\ %\cite{chu2019meansum} &
 \textcolor{blue}{These are great for the price. The tights are comfortable and don't take up much space.} \textcolor{red}{The only thing is that they can be worn to wear with the flip flops} $\dots$
 (I'm not sure if you have to wear them).
\\
\hline 
\textbf{Summary by Trace:}  \\  
\textcolor{blue}{These are great for those who want to wear a small. They are very comfortable and fit well.} The only problem is that \textcolor{red}{they don't last as long as some of the more expensive ones in the past.} \textcolor{blue}{I would recommend these to anyone.}
\\
\hline
\end{tabular}}
\caption{Example summaries generated by Coop~\cite{iso2021convex}, TRACE~\cite{zhang2023disentangling} for negative reviews. The \textcolor{red}{red} part represents negative, and the \textcolor{blue}{blue} is positive.}
\label{example}
\end{table}

However, as shown in Table~\ref{example}, the current opinion summarization approaches such as Coop and TRACE, are reluctant to generate a negative opinion summary given the input of negative opinions. 
We further conducted quantitative analysis and found that the emotional precision of the negative summaries generated by the current approaches is very limited, ranging from 10\% to 55\%.
Such significant sentiment bias might be attributed to the extremely unbalanced sentiment distribution in the dataset. Specifically, the proportion of reviews with a rating of more than 3 (positive) is 72.26$\%$ in the Yelp dataset, while 83.5$\%$ in the Amazon dataset. 

In the existing bias mitigation methods, modifying the data distribution can fundamentally eliminate bias and is not limited to specific model frameworks, exhibiting strong generalization capabilities ~\citep{dixon2018measuring, pruksachatkun2021does, qian2022perturbation}. Due to the powerful generation capabilities of large language models(LLMs), many works utilize them either as labelers to annotate unlabeled data 
\citep{yoo-etal-2021-gpt3mix-leveraging,wang-etal-2021-want-reduce}, or as generators to produce new data samples \citep{ye2022zerogen, meng2022generating, lm-cppf, gao2023self}. 
However, data augmentation based on LLMs has some drawbacks. 1) potentially risky. some studies have raised concerns about potential issues or toxicity in synthetic data from LLMs\cite{li2023large,li2023synthetic,pan2023risk}. 2) expensive cost. Balancing the emotional distribution of a million-scale dataset requires generating a large amount of synthetic data, making direct data generation using large models potentially expensive. 

%lass or lasso
Therefore, in the paper, we proposed LASS, a novel framework based on both \textbf{LA}rge and \textbf{S}mall language models for debia\textbf{S}ing opinion summarization. Firstly, a small size of synthesized negative reviews is obtained by rewriting the positive text via a large language model. We design prompts to ensure that the large pre-trained language model follows the minimal-edit principle when generating the counterfactual samples with opposite sentiments. Moreover, some counterfactual sample pairs of specific data sets are manually rewritten and used as samples inside the prompts for input into the generator to ensure that the modification of aspects and emotions is synchronized and reasonable.

Secondly, a disentangle reconstruction model is trained based on the generated data. Specifically, a disentangled autoencoder is proposed to obtain the sentiment and content representation through reconstruction, emotion, and distance constraints. Further, the new representations are obtained by exchanging the sentiment representation of the pair of counterfactual data, which are used to generate each other as the counterfactual reconstruction loss. To further constrain the emotion information, the original emotion representation is replaced with a learnable emotion label representation, where the weight depends on the outcome of emotion classification. 
Finally, a large amount of synthetic data can be obtained by decoding the new representation obtained from the combination of different sample representations and filtering based on confusion degree and sentiment classification.  

The experimental results demonstrate that LASS achieved results comparable to LLMs only, with an average reduction of 265,000 synthetic data points. 
Employing LASS for data augmentation across the three models resulted in an average increase of 36\% in negative sentiment accuracy without affecting the Rouge scores of the summaries, compared to 37\% with LLMs only.

The main contributions of this paper are as follows:
\begin{itemize}
\item We propose LASS, a data augmentation framework combining large and small language models to alleviate emotional bias by optimizing the emotional distribution of datasets. 
\item  We design a data reproduction method based on a disentangle reconstruction model, which generates additional data via decoding the combined new representations and filtering based on confusion degree and sentiment classification.  
\item The experimental results demonstrate that LASS which combines large and small models can alleviate sentiment bias as effectively as the approach solely based on LLMs, but more economically.

\end{itemize}

\section{Related Work}

\begin{figure*}[!t]
\centering
\includegraphics[width=0.85\textwidth]{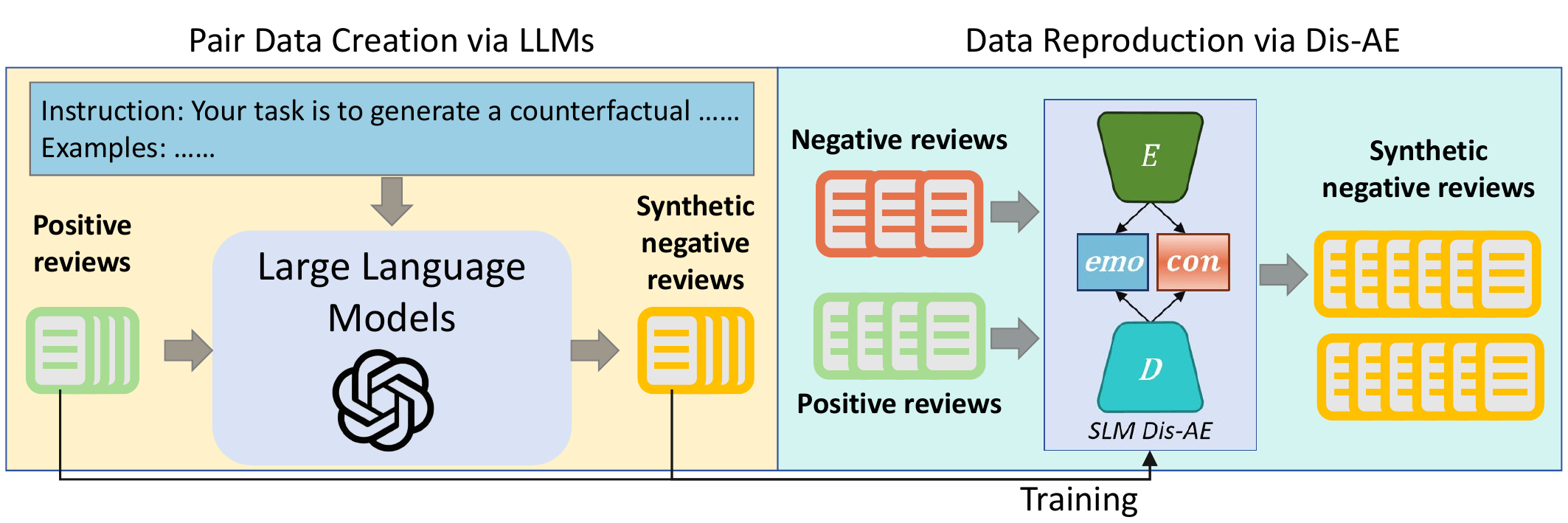} 
\caption{The architecture of LASS.}
\label{fig0}
\vspace{-0.2cm}
\end{figure*}

\subsection{Opinion Summarization}

Opinion summarization generally focuses on user reviews about products, hotels, restaurants, and so on. The abstractive approaches mainly utilize an encoder-decoder architecture, exploring various structures such as AE, VAE, or denoising autoencoder(DAE)\citep{chu2019meansum,copycat-bravzinskas2020unsupervised,dae-amplayo2020unsupervised,iso2021convex,zhang2023disentangling}. During training, these models are constrained by the objective of reconstructing the input text, and during generation, they use the average of text representations as the summary representation for decoding. Subsequent approaches aimed to enhance the controllability of generating summaries by explicitly \citep{suhara2020opiniondigest,elsahar2021self,amplayo2021aspect,ke2022consistsum} or implicitly \citep{amplayo2021unsupervised-content} modeling aspect information. 
Some methods also explore ways to fuse input information for summarization beyond simple averaging, utilizing techniques like composite optimization \citep{iso2021convex}, Wasserstein barycenter \citep{song2022unsupervised-wassos}, or hierarchical discrete latent space \citep{hosking2023attributable}.

\subsection{Debiasing Strategies in NLP}

Bias in NLP systems can typically be categorized as internal bias and external bias\citep{elsafoury2023bias,li2023survey}, depending on whether the bias is related to the training data of downstream tasks. Internal bias often pertains to issues of social fairness\citep{parraga2022debiasing}, such as gender and racial bias, which have been identified in the embeddings of pre-trained language models \citep{guo2022auto}. Existing work has attempted to address these issues through methods like adjusting pre-training data, introducing additional objectives, or post-processing.

On the other hand, external bias related to downstream tasks is often associated with task-specific features, such as entity bias in fake news detection \citep{zhu2022generalizing}, position bias in emotion cause extraction \citep{yan2021position}, and language bias in Visual Question Answering (VQA) \citep{cadene2019rubi}, and so on. To mitigate these specific biases, two distinct approaches have been developed: data distribution-related and model training-related \citep{shah2020predictive,parraga2022debiasing,li2023survey}.
In the data distribution-related approach, efforts are made to re-sample, weight, or generate data to counteract bias~\citep{dixon2018measuring,pruksachatkun2021does,qian2022perturbation}. In contrast, model training-related methods explore adversarial techniques, causality~\citep{cadene2019rubi,zhu2022generalizing}, disentanglement, and additional auxiliary modules to mitigate bias.

\section{Methodology}

In this section, we describe LASS, the data augmentation debias method via both LLMs and a small generator, a disentangle autoencoder. As Figure~\ref{fig0} shows, the overall architecture of LASS contains three processes, pair data creation via LLMs, Dis-AE model training, and data reproduction via Dis-AE. We first employ the LLMs with manual demonstrations to obtain pairs of counterfactual data. Then, based on pair samples, the disentanglement reconstruction model, Dis-AE, is elaborated in Section \ref{DMDA} with the training. Finally, large-scale negative reviews are generated by data reproduction based on Dis-AE.

% reproduction Hybridization

\subsection{Pair Data Creation via LLM}
\label{LMDA}

To avoid generating negative reviews that contain unreasonable product information, we obtain synthetic data by rewriting the original positive text. Adhering to the principle of minimal modification, synthetic data with the opposite sentiment but identical content is generated through LLMs via prompt with manual demonstration. Then the synthetic and the original form counterfactual data pairs, which are used to train the disentanglement generator.

\subsubsection{Prompt Design}

We first devised a foundational prompt to leverage the in-context learning capabilities of LLM for obtaining emotional opposite reviews. Then we enhance the prompt design by incorporating human-annotated samples and revising the order of examples in the prompts.

Formally, our foundational prompt is defined as a demonstration set $P$, comprising a task instruction $D$ and $k$ demonstration examples. Thus, we have $P = \{D, s(x_1, y_1), \cdots, s(x_k, y_k)\}$, where $s(x_i, y_i)$ denotes an pairwise example of emotional counterfactuals. Specifically, we define task instruction $D$ as \textit{"Your task is to generate a counterfactual that retains internal coherence and avoids unnecessary changes."} and randomly select $k$ samples from counterfactually-augmented movie reviews dataset \citep{Kaushik2020Learning}, where $k = 5$. Furthermore, we designate the temperature parameter as $T = 0.2$ to encourage a more deterministic output from the language model. 

The foundational prompts are already capable of enabling LLMs to flexibly generate counterfactuals, for example, when given the input \textit{"Jose's bandana must be giving him superpowers when he's cooking!!"}, the model generates the counterfactual as \textit{"maybe Jose's bandana is covering his eyes when he's cooking!!"}. 
However, there are still shortcomings in its performance, specifically manifested as incomplete transformations, where some positive text is retained, and illogical text (such as "disliking but frequently visiting").
Therefore, specific examples from corresponding datasets should be added to the data-specific prompt. We started with a collection of rewrite failure examples based on manual evaluations. Then a small evaluation dataset $\mathcal{I}$ is constructed via random selection, which consists of $m$ raw reviews with transformation issues and $n$ reviews corresponding to reasonable transformations. 

Afterward, we use an iterative approach to improve the prompt by observing the success rate of LLMs on the test set after adding manually annotated examples without any previous examples. Specifically, we randomly select review $x_t$ from set $\mathcal{I}$ and obtained example $s(x_t, y_t)$ manually. Then, we insert $s(x_t, y_t)$ into the current sequence of examples $\mathcal{C}$. The success rate of the LLMs in rewriting samples on the test set \{$\mathcal{I}$ - $\mathcal{C}$\} determines the best insertion position. Optimization stops when the improvement in success rate after adding new examples is less than $\varepsilon $, or when the overall success rate reaches $\delta$. The more detailed steps of the procedure and final prompt are in appendix \ref{Algorithm} and \ref{old_prompt}.

\begin{figure*}[!t]
\centering
\includegraphics[width=0.86\textwidth]{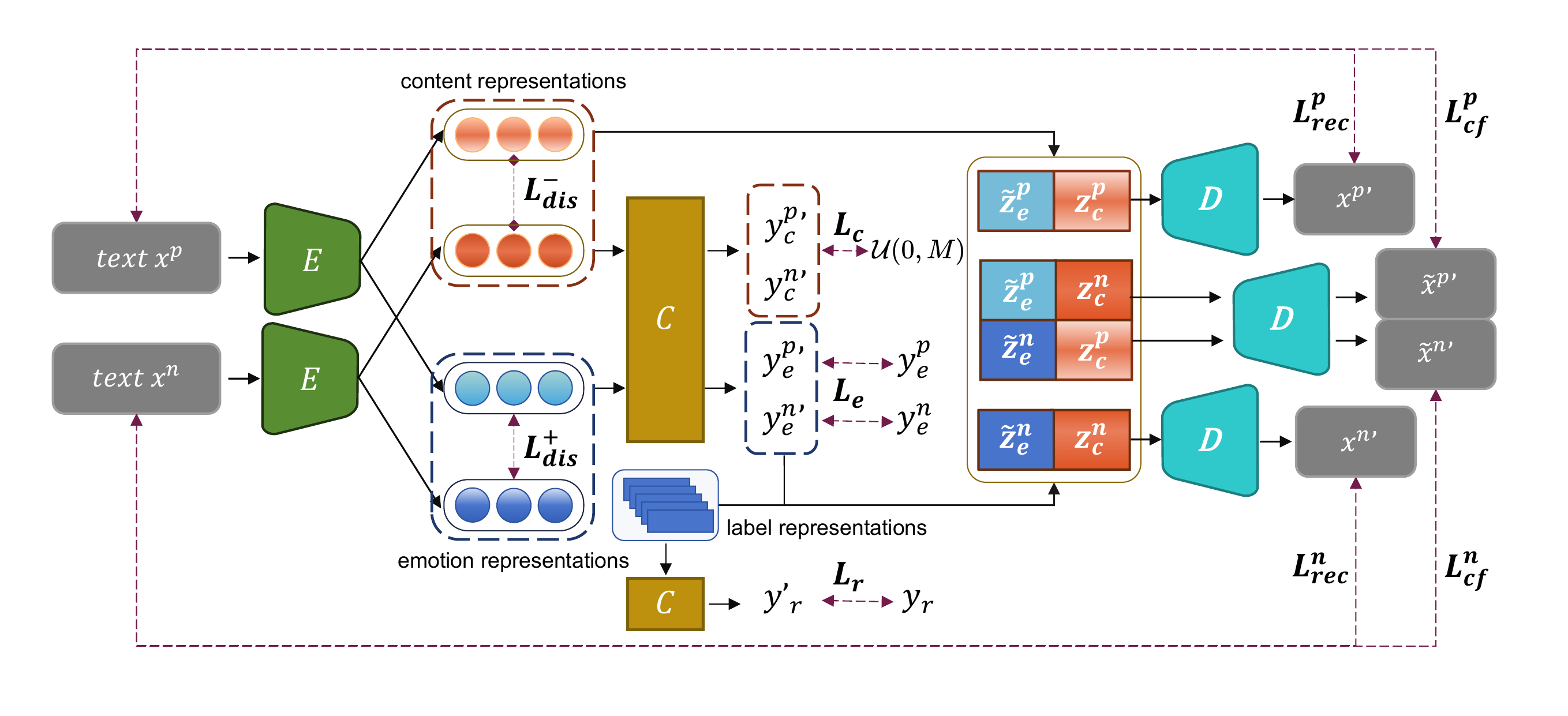} 
\caption{The architecture of disentanglement Model, Dis-AE. \textbf{$E$} and \textbf{$D$} are the encoder and the decoder. $y^p_e$, $y^n_e$ and $y_r$ are the emotion labels corresponding to the input $x^p$, $x^n$ and label representation. \textbf{$C$} is a sentiment classifier. $M$ is the number of categories for emotion classification.}
\label{fig1}
% \vspace{-0.2cm}
\end{figure*}

\subsection{Data Reproduction via Dis-AE}
\label{DMDA}

In this section, we describe data reproduction via the disentanglement reconstruction autoencoder, Dis-AE. We first describe the detailed components of the generator. Then, present the calculation process of Dis-AE and explain how to train it. Finally, we introduce the data enhancement process, Data Reproduction through Dis-AE.

% \subsubsection{Architecture Overview}

Given a set of text pairs (user reviews) with the same content but opposite emotional polarity, the aim of Dis-AE is to reconstruct the input pairs. As Figure~\ref{fig1} shows, the overall architecture of Dis-AE contains three components, an encoder $p_\theta$, an emotional classifier $C$, and a decoder $q_\phi$.

\noindent\textbf{The Encoder $p_\theta$} 
\citet{iso2021convex} show that large pre-training language models such as BERT \cite{kenton2019bert} and GPT-2 \cite{radfordlanguage-GPT} do not show a significant performance advantage over more lightweight model structures in unsupervised opinion summarization. Therefore, we employ the BIMEANVAE model~\cite{iso2021convex} which uses BiLSTM as encoder $p_\theta(z_e,z_n\mid x)$ and applies a mean pooling layer to the BiLSTM layer to obtain the primitive text representation $h$. Afterward, sentiment representation $z_e$ and content representation $z_n$ are obtained separately through different self-attention layers.

\noindent\textbf{The Emotional Classifier $M$} 
The sentiment vectors $z^p_e$ and $z^n_e$ and content vectors $z^p_c$ and $z^n_c$ are fed into classifier $C$ separately. The prediction result of emotion representation $z^p_e$ and $z^n_e$ should be corresponding labels $y_e$ and $y_n$, corresponding to 5 and 1  in the sentiment rating of the dataset. The prediction result of content representation does not contain sentiment information and should be uniform distribution $\mathcal{U}(0, M)$, where $M$ is the number of categories for emotion classification.

\noindent\textbf{The Decoder $q_\phi$} 
Following \citet{iso2021convex}, LSTM is employed as the decoder $q_\phi$. The distribution $q_\varphi(x\mid z)$ is computed by the reconstruction of the input $x^p$ from $z^p$ or $\widetilde{z}^p$. 

\subsubsection{Training of Dis-AE}

A pair of texts $x^p$ and $x^n$ is given which include consistent content and opposite emotional polarity $y_e$. In the training stage, the positive text $x^p$ is passed to the encoder $p_\theta(z_e,z_n\mid x)$ to get two types of text representation, the sentiment $z^p_e$ and the content $z^p_c$. Similarly, $z^n_e$ and $z^n_c$ can be obtained for $x^n$. Since the content of the paired texts is similar, but the emotion is opposite. Their content representations $z^p_c$ and $z^n_c$ are constrained to resemble each other, while their emotional representations $z^p_e$ and $z^n_e$ are forced to distance themselves.

These representations are all fed are put into the same emotional classifier $C$. 
To ensure that the emotion representation contains as little content information as possible, a learnable emotion label representation set $Z_r$ is used to replace $z^p_e$ and $z^n_e$. $Z_r$ also constrains by emotion classification loss $L_r$ and contains $M$ emotion label representations,  where $M$ is the number of categories for emotion classification. $M$ is the number of categories for emotion classification. Based on the emotion distribution $\hat{y}^p_e$ and $\hat{y}^n_e$ obtained by the corresponding emotion representation, the representation set $Z^r$ is weighted to get the final emotion representation $\widetilde{z}^p_e$ and $\widetilde{z}^n_e$.

Then the document latent variable $z^p$ is obtained by concatenating $\widetilde{z}^p_e$ and $z^p_c$, which is used to reconstruct the input text $x^p$ through the decoder $q_\phi(x \mid z)$. Since pairs of text have similar content representations, combining another content representation $z^e_c$ should also represent the current text $x^p$. Thus, positive counterfactual representations $\widetilde{z}^p$ are obtained by a combination of $\widetilde{z}^p_e$ and $z^n_c$, which is decoded to obtain $x^p$. Similarly, $\widetilde{z}^n_e$ is combined separately with $z^n_c$ and  $z^p_c$, and decoded to obtain $x^c$.

In order to ensure the basic ability of text generation, we retained the AE constraints, the reconstruction loss $L_{rec}$. When reconstructing the input pair separately, representation $z^p$ from concatenated $z^p_e$ and $z^p_c$ is used as the input of the decoder to reconstruct the input text $x^p$. The same procedure is applied to obtain the corresponding negative text $x^n$. The reconstruction loss is defined as:

\begin{equation}
\begin{aligned}
& L_{rec}(\theta, \phi)  =  \\
& -\sum_{i=1}^{N}\underset{p_{\theta}\left(\widetilde{z}^p_e,z^p_c\mid x^p\right)}
{\mathbb{E}}[\log q_{\phi}(x^p\mid \widetilde{z}^p_e,z^p_c)] \\
& -\sum_{i=1}^{N}\underset{p_{\theta}\left(\widetilde{z}^n_e,z^n_c\mid x^n\right)}
{\mathbb{E}}[\log q_{\phi}(x^n\mid \widetilde{z}^n_e,z^n_c)],
\end{aligned}
\end{equation}
where $\theta$ and $\phi$ are the parameters of the model. The reconstruction loss improves the quality of the decoded text and forces the text representation to store content information with emotion. 
To disentangle emotional representation and content representation,  we employ an emotional auxiliary constrain $\mathcal{L}_{emo} = L_{e} + L_{n} + L_{r}$, which is including with emotion classification constraints $L_{e}$, emotion adversarial constraints $L_{c}$ and label emotion constraints $L_{r}$.

The sentiment representation $z^p_e$ and $z^n_e$ and content representation $z^p_c$ and $z^n_c$ are fed into classifier $C$ separately. The prediction result of $z^p_e$ and $z^n_e$ should be the corresponding emotion label $y_e$ and $y_n$, which is a cross-entropy loss:
\begin{equation}
\begin{aligned} 
L_{e}(\theta) = & -{\mathbb{E}_{p_{\theta}\left(z^p_e\right)}} \sum_{i=1}^{M} y^p_e log (p(\hat{y}^p_e|z^p_e)) \\
& -{\mathbb{E}_{p_{\theta}\left(z^n_e\right)}} \sum_{i=1}^{M} y^n_e log (p(\hat{y}^n_e|z^n_e)).
\end{aligned}
\label{e}
\end{equation}

Inspired by \citet{pergola2021disentangled}, rather than being unable to achieve correct classification, we assume that content representations $z^p_c$ and $z^n_c$ are sentiment-neutral, and should not exhibit any category bias during sentiment classification. Therefore, $z^p_c$ and $z^n_c$ should be fed into the sentiment classifier to obtain a uniform sentiment classification distribution, which is an expected KL divergence loss:  

\begin{equation}
\begin{aligned} 
 & L_{n}(\theta) = -{\mathbb{E}_{p_{\theta}\left(z^p_c\right)}} [\mathbb{D}_{KL}(\mathcal{U}(0,M) || p(\hat{y}^p_c|z^p_c))] \\
 & -{\mathbb{E}_{p_{\theta}\left(z^n_c\right)}} [\mathbb{D}_{KL}(\mathcal{U}(0,M) || p(\hat{y}^n_c|z^n_c))],
\end{aligned}
\end{equation}
where $M$ is the total number of sentiment classes. The former is the expected KL divergence with the uniform distribution $\mathcal{U} (0, M)$. Given that an
additional learnable label representation set $Z^r = \{z^r_1, \cdots, z^r_M\}$ is used to replace the emotion representations $z^p_e$ and $z^n_e$, $Z^r$ also need to contain emotional information constrained by a similar loss of emotional classification:

\begin{equation}
\begin{aligned} 
L_{r} = & -\sum_{i=1}^{M} y^r_i log (p(\hat{y}^r_i|z^r_i)).
\end{aligned}
\label{r}
\end{equation}

To further introduce relational knowledge hidden in pairs of data, we add distance loss $\mathcal{L}_{dis}$ and counterfactual reconstruction loss $\mathcal{L}_{cf}$. The distance loss is based on the prior knowledge that the input text pair expresses opposite emotions but shares similar content. The represented distance is constrained based on the sentence similarity:

\begin{equation}
\begin{aligned} 
& L_{dis} = 2 + sim(z^p_e, z^n_e) - sim(z^p_c, z^n_c),
\end{aligned}
\end{equation}
where $sim(\cdot)$ indicates the cosine similarity function. Likewise, since the text pair $x^p$ and $x^n$ contain the same content information, the alternate content representation should allow for successful decoding of the corresponding text. Thus the counterfactual reconstruction loss is: 

\begin{equation}
\begin{aligned}
& L_{cf}(\theta, \phi)  =  \\
& -\sum_{i=1}^{N}\underset{p_{\theta}\left(\widetilde{z}^p_e,z^p_c\mid x^p\right) p_{\theta}\left(\widetilde{z}^n_e,z^n_c\mid x^n\right)}
{\mathbb{E}}[\log q_{\phi}(x^p\mid \widetilde{z}^p_e,z^n_c)] \\
& -\sum_{i=1}^{N}\underset{p_{\theta}\left(\widetilde{z}^n_e,z^n_c\mid x^n\right) p_{\theta}\left(\widetilde{z}^p_e,z^p_c\mid x^p\right)}
{\mathbb{E}}[\log q_{\phi}(x^n\mid \widetilde{z}^n_e,z^p_c)].
\end{aligned}
\end{equation}

Our final objective function is:
\begin{equation}
\mathcal{L} = L_{rec} + \alpha\mathcal{L}_{emo} + \beta\mathcal{L}_{dis} + \gamma \mathcal{L}_{cf} ,
\end{equation}
where $\alpha$, $\beta$ and $\gamma$ are hyper-parameters that controls the strength of constrains. 

\subsubsection{Data Reproduction}

After training, data reproduction can be performed by selecting parent samples from the training set and combining them with the disentanglement model Dis-AE. Specifically, when negative reviews for a specific product are needed, positive reviews for that product are selected along with any negative reviews as parents. The parent samples are inputted into Dis-AE to obtain sentiment representations and content representations separately. By combining the content representation of positive reviews with the sentiment representation of negative reviews, we obtain the child representation. Decoding the child representation yields negative samples. This data reproduction approach ensures the controllability of content and sentiment of generated text while also meeting the demand for large-scale data augmentation, due to the diversity of parental sample combinations.

Due to the limitation of small model generation ability, the generated text may be unreadable, or with incorrect sentiment polarity. Therefore, we add a data filtering process based on perplexity and sentiment classification to ensure the quality of the generated text.

\section{Experiments} 

\subsection{Datasets}

We performed experiments on two opinion summarization
benchmarks, the Amazon dataset \citep{copycat-bravzinskas2020unsupervised} and Yelp \cite{chu2019meansum}. All datasets include review ratings with a 1–5 scale which we used as sentiment labels. Besides training reviews, these two datasets also contain gold-standard summaries for 200 and 60 sampled objects for evaluation. 

However, extreme sentiment biases also exist in the evaluation data. Therefore, we extracted 800 positive and 800 negative products from the training data of both datasets. Half for the validation, and the other half for the test. Each product consists of 7 or 8 reviews, all rated as 5 for positive or 1 for negative sentiment. Due to the consistent sentiment polarity of reviews, we utilized them for assessing the ability of summary generation to produce summaries with different sentiment polarities for positive (POS) and negative products (NEG).

\begingroup
\renewcommand{\arraystretch}{1.10}
\begin{table*}[!th]
\centering
\scalebox{0.80}{
\begin{tabular}{l||ccc|ccc||ccc|ccc}
\hline 
 & \multicolumn{6}{c||}{Amazon} & \multicolumn{6}{c}{Yelp} \\ 
 \hline
& \multicolumn{3}{c|}{Pos} & \multicolumn{3}{c||}{Neg} & \multicolumn{3}{c|}{Pos} & \multicolumn{3}{c}{Neg} \\
 (\%) & Rev & Sen & Dif & Rev & Sen & Dif & Rev & Sen & Dif & Rev & Sen & Dif \\
 \hline 
 %MeanSum & $28.27$ & $3.54$ \\ 

 Wassos(T) & $93.25$ & $88.97$ & - & $20.63$ & $19.84$ & - & $98.25 $ & $91.51$ & - & $43.5$ & $47.25$ & -\\ 
 Wassos(O) & $93.5$ & $92.49$ & -& $7.13$ & $10.31$ & -& $79.25$ & $78.93$ & -& $59.25$ & $53.28$ & -\\ 
 TRACE(a) & $91.63$ & $82.29$ & -& $24.38$ & $29.61$ & -& $100$ & $94.53$ & -& $68.5$ & $57.08$& - \\
 TRACE & $89.25$ & $80.94$ & -& $40.5$ & $38.82$ & -& $99.5$ & $97.44$ & -& $8.5$ & $10.92$ & -\\	

 \hline
 Copycat & $\textbf{93.75}$ & $\textbf{84.69}$ & -& $16.25$ & $16.40$ & -& $\textbf{97.75}$ & $\textbf{88.43}$ & -& $47.75$ & $41.15$ & -\\
 \quad +GPT  & $60.95$ & $57.00$ & $-30.3$ & $70.63$ & $55.09$ & $+46.5$ & $95.00$ & $76.71$ & $\textbf{-7.2}$ & $78.13$ & $63.96$ & $+26.6$  \\
 \quad +LASS  & $61.13$ & $64.01$ & $\textbf{-26.7}$ & $\textbf{76.75}$ & $\textbf{58.34}$ & $\textbf{+51.2}$ & $93.38$ & $76$ & $-8.4$ & $\textbf{86.5}$ & $\textbf{64.02}$ & $\textbf{+30.8}$\\
  \hline
 Coop(a) & $81.75$ & $76.05$ & - & $46.88$ & $41.39$ & - & $99.875$ & $93.23$ & - & $34$ & $39.31$ & - \\
\quad +GPT  & $\textbf{94.38}$ & $88.26$ & $\textbf{+12.4}$ & $\textbf{90.88}$ & $\textbf{79.68}$ & $\textbf{+41.1}$ & $99.63$ & $94.98$ & $+0.7$ & $77.50$ & $\textbf{73.48}$ & $+38.8$  \\
 \quad +LASS  & $92.75$ & $\textbf{86.42}$ & $+10.7$ & $89.38$ & $74.14$ & $+37.6$ & $\textbf{99.75}$ & $\textbf{97.26}$ & $\textbf{+2.0}$ & $\textbf{79.25}$ & $72.37$ & $\textbf{+39.2}$  \\
  \hline
 Coop &$82.75$ & $76.97$ & - & $58$ & $47.64$ & - & $99$ & $92.38$ & - & $51.5$ & $47.55$ & - \\
 \quad +GPT  & $ 90.63$ & $81.40$ & $+6.2$ & $\textbf{93}$ & $\textbf{76.48}$ & $\textbf{+31.9}$ & $\textbf{100}$ & $\textbf{95.52}$ & $\textbf{+2.1}$ & $\textbf{93.38}$ & $\textbf{80.86}$ & $\textbf{+37.6}$  \\
 \quad +LASS  & $\textbf{90.75}$ & $\textbf{82.38}$ & $\textbf{+6.7}$ & $84.88$ & $68.89$ & $+24.1$ & $99.75$ & $95.41$ & $+1.9$ & $90.25$ & $77.13$ & $+34.2$  \\

\hline
\end{tabular}
}

\caption{Sentiment accuracy results on Amazon and Yelp. The bold scores denote the best scores. On Amazon, The amount of data enhanced via GPT for Copycat, Coop (a), and Coop are 450k, 360k, and 360k respectively, while on Yelp, they are 630k, 450, and 540k respectively. The synthetic data used to train the Dis-AE model in LASS is 200k, consistent across all models and datasets.
%\dag from \cite{songunsupervised-wassos}.
}
\label{result1_e}
\vspace{-0.2cm}
\end{table*}
\endgroup

\begingroup
\renewcommand{\arraystretch}{1.10}
\begin{table}[!th]
\centering
\scalebox{0.80}{
\begin{tabular}{l||ccc||ccc}
\hline 
 & \multicolumn{3}{c||}{Amazon} & \multicolumn{3}{c}{Yelp} \\ 
 \hline
 & \textbf{R1} & \textbf{R2} & \textbf{RL} & \textbf{R1} & \textbf{R2} & \textbf{RL} \\
 \hline 
 %MeanSum & $28.27$ & $3.54$ \\ 
 Wassos(T) & $29.7$ & $6.5$ & $20.0$ & $30.8$ & $5.9$ & $18.3$\\ 
 Wassos(O) & $32.5$ & $7.2$ & $21.8$ & $26.6 $ & $4.5$ & $16.4$ \\ 
 TRACE(a)  & $33.7$ & $6.3$ & $20.5$ & $32.6$ & $6.6$ & $20.0$ \\
 TRACE   & $36.0$ & $7.2$ & $20.8$ & $33.9$ & $6.8$ & $19.7$ \\

 \hline
 Copycat & $31.9$ & $\textbf{6.1}$ & $\textbf{20.4}$ & $29.3$ & $5.4 $ & $17.7$ \\
 \quad +GPT & $\textbf{32.3}$ & $5.9$ & $19.7$ & $\textbf{30.0}$ & $5.6$ & $18.8$ \\
 \quad +LASS  & $31.8$ & $5.8$ & $19.5$ & $29.4$ & $\textbf{6.0}$ & $\textbf{19.2}$ \\
  \hline
 Coop(a) & $32.1$ & $5.1$ & $18.1$ & $30.6$ & $5.9$ & $18.8$\\
 \quad +GPT & $32.2$ & $\textbf{7.1}$ & $20.2$ & $\textbf{31.6}$ & $6.4$ & $19.5$ \\
 \quad +LASS & $\textbf{32.9}$ & $6.3$ & $\textbf{20.4}$ & $31.6$ & $\textbf{6.9}$ & $\textbf{19.6}$ \\
  \hline
 Coop    & $35.7$ & $6.2$ & $19.8$ & $\textbf{34.5}$ & $\textbf{6.9}$ & $\textbf{19.6}$ \\
 \quad +GPT & $35.6$ & $6.4$ & $20.6$ & $34.0$ & $6.8$ & $19.5$ \\
 \quad +LASS & $\textbf{36.2}$ & $\textbf{7.0}$ & $\textbf{21.4}$ & $33.8$ & $6.9$ & $19.4$ \\
\hline
\end{tabular}
}
\caption{Rouge scores on Amazon and Yelp. The bold scores denote the best scores.
%\dag from \cite{songunsupervised-wassos}.
}
\label{result1_r}
\vspace{-0.2cm}
\end{table}
\endgroup

\subsection{Evaluation Metrics and Baselines}

We evaluate summary systems with the classical ROUGE-{1, 2, L} metrics \citep{lin2004rouge}. We also report sentiment precision about the positive and the negative at the sentence level (Sen) and review level (Rev), using the sentiment analysis model from Stanza \citep{qi2020stanza} to compute. All ratings are normalized to scores between 0 and 1. More details are in the Appendix.
The term "Dif" represents the average change in sentiment accuracy at both the review and sentence levels after data augmentation using GPT or LASS.

Following prior work \cite{iso2021convex,song2022unsupervised-wassos}, we compare with \textbf{Copycat}~\cite{copycat-bravzinskas2020unsupervised}, \textbf{Coop}~\cite{iso2021convex}, \textbf{Wassos}~\cite{song2022unsupervised-wassos} and TRACE \cite{zhang2023disentangling}. (a), (O), and (T) represent different clustering strategies for the model. The detailed introduction is in the Appendix.
Considering the sensitivity of the counter-templates in TRACE to training data, we experimented with data augmentation methods based on Chatgpt and LASS on three models Coop, Coop(a), and Copycat.

\subsection{Implementation Details}

In this work, we employ the ChatGPT platform \footnote{https://chat.openai.com/chat} to generate pairwise emotional counterfactuals within a crafted prompt setting. 
For the prompt optimization, $m = 40$, $n = 10$, $\delta = 80\%$ and $\varepsilon = 10\%$.The final prompts include 5 pairs of examples for the Amazon dataset and 7 pairs for Yelp. Specifically, we extract the samples with a sentiment score of 5 from the training data. 

\begin{figure*}[!t]
\centering
\includegraphics[width=1.0\textwidth]{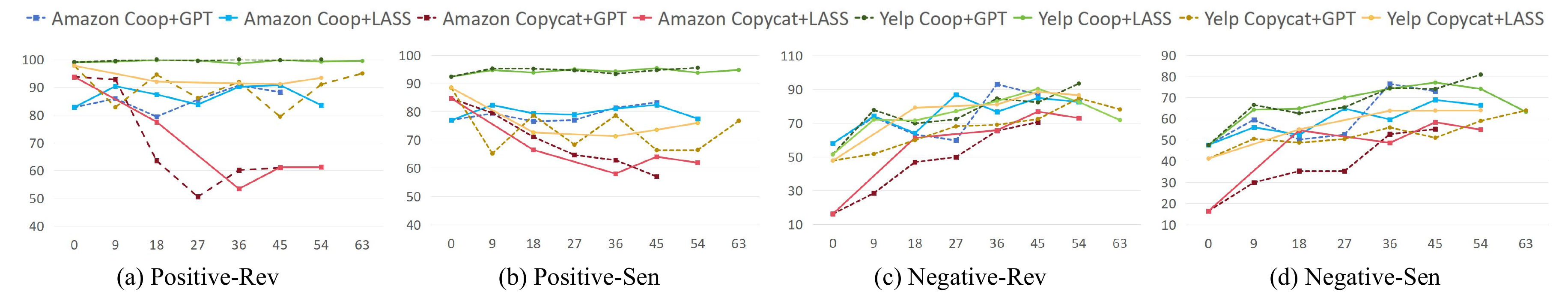} 
\caption{Experimental results about Coop and Copycat with different sizes of synthesized data from GPT or LASS on Amazon and Yelp. The horizontal axis represents the amount of synthetic data added, measured in increments of 10k. For example, 9 corresponds to 90k.}
\label{fig3}
\vspace{-0.2cm}
\end{figure*}

For the disentanglement model Dis-AE, we used Adam optimizer ~\cite{kingma2015adam} with a linear scheduler, whose initial learning rate is set to $5e^{-4}$. For beam search in the generation, the beam size is set to 4 and a max token size of 70. The amount of training data used is 200k, according to the analysis in Section \ref{analysis}. Additionally, based on the PPL testing conducted on the training set, we set the threshold for PPL at 125. Only generated samples with PPL less than 125 and classified as negative by review level sentiment classifier from Stanza \citep{qi2020stanza} are retained.
To prevent the imbalance of multiple constraints from undermining the text generation capability, we mimic KL annealing~\cite{li2019surprisingly,iso2021convex} to gradually increase $\alpha$, $\beta$, and $\gamma$ from 0 during training. The upper limit for the weight of sentiment loss $\alpha$ is set to 5, while $\beta$ and $\gamma$ are both limited to 1. All experiments were conducted on NVIDIA GeForce RTX 3090 or NVIDIA Tesla V100. 

\subsection{Results}

According to table \ref{result1_e}, Synthesized data from both LASS and GPT significantly enhance the model's performance in nearly all sentiment accuracy measures, whether at the review or sentence level. 
The exception is the copycat model, despite improving negative sentiment accuracy, harms positive sentiment accuracy when augmenting negative sentiment data. However, comparatively, LASS improved negative sentiment accuracy more than GPT. And LASS maintains a higher positive sentiment accuracy for the Amazon dataset.

This kind of exception may be attributed to the multiple influences of data augmentation methods, summarization models, and datasets. From the perspective of summarization models, the overall performance of the Copycat model is inferior to that of Coop(a) and Coop in terms of both sentiment accuracy and ROUGE scores. For positive sentiment accuracy, any model's performance on the Yelp dataset as a whole is better than that on Amazon. This may be because the Yelp data mainly consists of restaurant reviews, making it easier for models to learn expressions of positivity and negativity compared to the diverse product types in the Amazon data.

From the ROUGE scores in Table \ref{result1_r}, it was observed that all methods did not exhibit a performance decrease after data augmentation using GPT or LASS. This suggests that the data augmentation methods are applicable across different models and do not degrade the performance of the models on the original task. It also indicates that both the GPT and LASS methods generate highly readable data, and even with a large-scale addition to the training data, they do not disrupt the training of summarization tasks.

\begingroup
\renewcommand{\arraystretch}{1.10}
\begin{table}[!t]
\centering
\scalebox{0.85}{
\begin{tabular}{c|cccc}
\hline 
 $Num(k)$ & $\textbf{PPL}\downarrow$  & $\textbf{R1}$ & $\textbf{R2}$ & $\textbf{RL}$ \\
\hline 
 $50$ & $540.25 $ & $53.61 $ & $16.80 $ & $34.74 $ \\
 $100$ & $1314.50 $ & $60.57 $ & $23.31 $ & $42.65 $ \\
 $150$ & $788.96 $ & $56.25 $ & $34.08 $ & $50.92 $  \\ 
 $200$ & $360.13 $ & $59.55 $ & $38.85 $ & $54.86 $ \\ 
 $250$ & $403.57 $ & $59.29 $ & $39.36 $ & $54.44$  \\
\hline
\end{tabular}}
\caption{Experimental results about Dis-AE with different sizes of train data on Amazon. 
%\dag from \cite{songunsupervised-wassos}.
}
\label{result_number}
\vspace{-0.2cm}
\end{table}
\endgroup

\subsection{Analysis}
\label{analysis}

To investigate the impact of different synthetic data on summarization models, we analyzed the sentiment accuracies of different summarization models using varying amounts of augmentation data from GPT or LASS, as shown in Figure \ref{fig3}. 
Overall, adding negative reviews can improve the negative sentiment accuracy of summaries, while may affect the ability to generate positive summaries to some extent. For Coop, the positive accuracy on Amazon shows some instability as the data volume increases. Meanwhile, Copycat's positive accuracy experiences a significant decline, suggesting that Copycat may not handle sentiment information well in summaries and tends to generate neutral text with mixed positive and negative sentiments.

Additionally, we explored the amount of data required for training Dis-AE. Evaluating whether the quality of the generated text meets the training requirements of summarization requires a lot of downstream experiments. To more efficiently confirm the data requirements, we employ two metrics: perplexity (PPL) and counterfactual reconstruction ROUGE score. The counterfactual reconstruction ROUGE score is similar to the counterfactual reconstruction loss $ L_{cf}$, calculating the ROUGE score of reconstructed text after exchanging paired counterfactual samples with target text. PPL relies on GPT-2 to compute the degree of text fluency \footnote{https://huggingface.co/docs/transformers/perplexity}.

The results, as shown in the table \ref{result_number}, indicate that the quality of generation improves steadily with the increase in data volume, with instabilities observed after reaching 200k. The reason why the PPL for 50k is less than that for 100k is because samples shorter than 10 characters are not included in the PPL calculation, as PPL becomes erratic for excessively short texts.

\section{Limitation} 

Overall, while debias through data augmentation can generalize across different models, its effectiveness is also limited by the performance and characteristics of each model. For example, in the current scenario, the Copycat model experienced significant degradation in positive sentiment accuracy after using augmented data on the Amazon dataset. For another model TRACE, changes in data distribution significantly affect the performance of the summaries, as observed in our preliminary experiments. This may be attributed to one of the parameters, the counter-template, being sensitive to the training data. Additionally, determining the minimum data required for Dis-AE training is a critical issue. The current approach, based on perplexity and counterfactual reconstruction metrics, only indirectly reflects the quality of generated counterfactual texts. We will continue to explore the training data requirement for Dis-AE in future work.

\section{Acknowledgements}
We would like to thank anonymous reviewers for their valuable comments and helpful suggestions. The authors acknowledge financial support from the National Natural Science Foundation of China (62176053). This research work is also supported by the Big Data Computing Center of Southeast University. YH was supported by a Turing AI Fellowship (EP/V020579/1, EP/V020579/2) funded by the UK Research and Innovation. 

\bibliography{custom}

\appendix

\section{Sentiment Evaluation}
\label{sec:appendix}

For positive reviews, the sentiment score is 1 for positive, 0.5 for neutral, and 0 for negative, while for the negative set, the negative is 1. The rating for the review level precision involves assigning a score to the entire text, while at the sentence level, scores are assigned to each sentence in the text and then averaged. 

\section{Baselines}

We compare our method against the following unsupervised summarization approach.
Copycat \citep{copycat-bravzinskas2020unsupervised} captures the dependency relationship between the product and reviews by defining a hierarchical VAE.
Coop \citep{iso2021convex} searches input combinations for the summary aggregation using the input-output word overlapping. \(a\) represents the use of a simple averaging strategy, while the other represents the retrieval strategy of Coop. 
Wassos \citep{song2022unsupervised-wassos} uses the Wasserstein barycenter of the semantic and syntactic distributions to obtain the summary.  \(O\) and \(T\) represent different clustering strategies.
TRACE \cite{zhang2023disentangling} is based on text representation disentanglement with generated counter-templates. \(a\) represents the use of a simple averaging strategy, while the other represents the retrieval strategy of Coop. 

\section{Algorithm}
\label{Algorithm}

\begin{algorithm}[!t]
\caption{Prompt Optimization} 
\label{al1}
\begin{algorithmic}[1]
\REQUIRE 
    instruction $D$,
    test set $\mathcal{I} = \{x_1, \cdots, x_{|\mathcal{I}|}\}$, 
    example permutation $\mathcal{S}$, 
    candidate example set $\mathcal{C} = \mathcal{I}$, 
    time step $t = 1$.
\ENSURE Optimized Prompt $P \leftarrow P_t$.
\REPEAT
    \STATE randomly select review $x_t$ from set $\mathcal{C}$ and obtained example $s(x_t, y_t)$ manualy.
    \STATE Insert $s(x_t, y_t)$  into $\mathcal{S}$ to earned permutation set $\{\mathcal{S}_t^1, \cdots, \mathcal{S}_t^{|s|+1}\}$, which each permutation contain $|\mathcal{S}| + 1$ examples.
    \FOR { $i = 1$ to $|\mathcal{S}| + 1$}
        \STATE $P_t^i = \{D, \mathcal{S}_t^i \}$;
        \STATE $score_t^i \leftarrow score(\{\mathcal{I}-\mathcal{S}\}|P_t^i)$;
    \ENDFOR
    \STATE update permutation $\mathcal{S}$: $\mathcal{S} = \underset{\mathcal{S}_t^i}{argmax}$ \space $score_t^i$;
    \STATE $\mathcal{C} = \{\}$;
    \STATE add $x_i$ into $\mathcal{C}$ if $score(x_i|P_t) < 0$;
    \STATE $t = t + 1$;
\UNTIL $score(\{\mathcal{I}-\mathcal{S}\}|P_t) > \delta$ or \\ $score(\{\mathcal{I}-\mathcal{S}\}|P_t) - score(\{\mathcal{I}-\mathcal{S}\}|P_{t-1}) < \varepsilon $.

\end{algorithmic}
\end{algorithm}

the success rate of the LLMs $score(S|P_t)$ indicates a score evaluating on dataset $S = \{ x_1, \cdots, x_k\}$ under prompt $P_t$, which defined as:
\begin{equation}
    score(S|P_t) = \sum_{i=1}^{|S|}HumanEval(LLM(x_i, P_t)),
\end{equation}
where $LLM(x_i, P_t)$ is LLM's output given input $x_i$ and prompt $P_t$. $HumanEval$ is a score given by human evaluation, whose value belongs to $\{ 0, 1 \}$, 1 demonstrates conformity to normative standards, and 0 indicates the issues in reasonableness or sentiment polarity after generation. 

\onecolumn
\section {Prompt}\label{old_prompt}
Here is the foundational prompt employed to obtain annotated validation datasets for prompt optimization:
\begin{framed}
Your task is to generate a counterfactual that retains internal coherence and avoids unnecessary changes.
\newline

Example: Really good movie. Maybe the best I've ever seen. Alien invasion, a la The Blob, with crazy good acting. Meteorite turns beautiful woman into a host body for nasty tongue. Engaging plot, great tongue. Absurd comedy worth watching. Maybe don't wash your hair or take out the trash but take time out to watch this movie.

Counterfactual: Really bad movie. Maybe the worst I've ever seen. Alien invasion, a la The Blob, without the acting. Meteorite turns beautiful woman into a host body for nasty tongue. Bad plot, bad fake tongue. Absurd comedy worth missing. Wash your hair or take out the trash.
\newline

Example: I rated this a 5. The dubbing was as good as I have seen. The plot - wow. I'm not sure which made the movie more great. Jet Li is definitely a great martial artist, as good as Jackie Chan.

Counterfactual: I rated this a 3. The dubbing was as bad as I have seen. The plot - yuck. I'm not sure which ruined the movie more. Jet Li is definitely a great martial artist, but I'll stick to Jackie Chan movies until somebody tells me Jet's English is up to par.
\newline

Example: Greenaway seems to have a habit of trying hard to entertain his viewers. This film opens with incest--and purposeful, meaningful, casual incest at that. That's Greenaway's focus. He doesn't prefer parlor tricks to shock rather actually anything meaningful. Technical skill isn't enough. He's a bit perverse for the sake of perversity but it works out well.

Counterfactual: Greenaway seems to have a habit of trying deliberately to disgust his viewers. This film opens with incest--and purposeless, meaningless, casual incest at that. That's Greenaway's big problem. He prefers parlor tricks to shock over actually doing anything meaningful. Technical skill isn't enough. He's just a bit perverse for the sake of perversity.
\newline

Example: This is one of the most awesome movies ever. Shaq better do more movies. This movie just gave me a good bit of life and I will always remember that. I will never make fun of this movie until I die, and then even after! It is just so wonderful and even funny. MST3000 would have a blast with this one.

Counterfactual: This is one of the most god-awful movies ever. Shaq better just stick to basketball. This movie took away apart of my life I will never have back. I will make fun of this movie until I die, and then some. It is so horrible it is not even funny. MST3000 would have a blast with this one.
\newline

Example: There's something wonderful about the fact that a movie made in 1934 can be head and shoulders above every Tarzan movie that followed it, including the bloated and boring 1980s piece Greystoke. Once the viewer gets past the first three scenes, which are admittedly dull, Tarzan and his Mate takes off like a shot, offering non-stop action, humor, and romance. Maureen O'Sullivan is charming and beautiful as Jane and walks off with the movie. Weismuller is solid as well. Highly recommended.

Counterfactual: There's something awful about the fact that a movie made in 1934 can be head and shoulders below every Tarzan movie that followed it, including the bloated and boring 1980s piece Greystoke. Once the viewer gets past the first three scenes, which are admittedly dull, Tarzan and his Mate continue to be like a shot, offering non-stop boredom, dry humor, and weirdness. Maureen O'Sullivan is mean and ugly as Jane and walks off with the movie. Weismuller is rude as well. Not recommended.
\end{framed}

\subsection{Added Examples After Prompt Optimization}\label{new_prompt}

In Prompt Optimization, we annotated $k_1$ examples from the Amazon dataset and $k_2$ examples from the Yelp dataset to gain better performance in the counterfactual generation, where $k_1 = 5$ and $k_2 = 7$.

Here are the annotated examples from the Amazon dataset:
\begin{framed}
Example: I tried connecting my iPhone 4S to my 2012 Ford Focus using a standard 3.5mm audio cable, but it sounded awful and noisy. Instead, I purchased this cable and now the audio going into my car sounds perfect! This is the best \$3-5 I could have spent to improve my car audio.

Counterfactual: I tried connecting my iPhone 4S to my 2012 Ford Focus using a standard 3.5mm audio cable, but it sounded awful and noisy. Instead, I purchased this cable and now the audio going into my car still sounds awful! This is the worst \$3-5 I could have spent to improve my car audio.
\newline

Example: I ordered this for my 3 yr old for Halloween. He loved it!! The candy catcher in the front is really neat, but probably need to take a pail or something else along also because it can get to be heavy if they get a lot of candy. I was very pleased with the way it fit and everything.

Counterfactual: I ordered this for my 3 yr old for Halloween. He prefer another one!! The candy catcher in the front is really small, but probably need to take a pail or something else along also because it can get to be heavy if they get a lot of candy. I was concerned about the way it fit and everything.
\newline 

Example: I loved this steamer when I got it, and it has remained a very stable item to use. I feel confident taking it out of the microwave when hot because it has never dumped hot food all over me.

Counterfactual: I disliked this steamer when I got it, and it has remained a very unstable item to use. I feel hesitant taking it out of the microwave when hot because it has frequently spilled hot food all over me.
\newline 

Example: Purse looks great. The bag is cute and flashy but the size is smaller than expected overall. The stones and straps are not very durable and break or fall off easily.

Counterfactual: The purse looks awful. The bag is unattractive and plain but the size is just the expected overall. The stones and straps are just durable and break or fall off not easily.
\newline

Example: The tank fit very well and was comfortable to wear. The material was thicker than I expected, and I felt it was a great value for the price. I've bought similar quality tanks for \$10 at a local store.

Counterfactual: The tank didn't fit well at all and it was quite uncomfortable to wear. The material was much thinner than I expected, and I felt it was not a good value for the price. I've bought similar quality tanks for less than \$10 at a local store.
\end{framed}

Here are the annotated examples from the Yelp dataset:
\begin{framed}
Example: Nothing special here. The music is too loud, the drinks too pricey, and the servers to shapely for the clothing they are wearing. Not that there are many options around job.com arena to choose from, sadly this is probably the best.

Counterfactual: A special place here. The music is just the right volume, the drinks are reasonably priced, and the servers are dressed decently. There are many good options around job.com arena to choose from, luckily this is probably the best.
\newline

Example: My wife and I had dinner and wine here during their last week open. The food and wine was fantastic as always. It is unfortunate that Twisted Rose closed its doors. They will be missed.

Counterfactual: My wife and I had dinner and wine here during their last week open. The food and wine was terrible as always. It is fortunate that Twisted Rose closed its doors. They will not be missed.
\newline

Example: Pro: Brightly lit, open late Con: Waaay overpriced unless you typically drive in the mud and need lots of car washes for a monthly fee.

Counterfactual: Con: Dimly lit, open early Pro: Surprisingly affordable unless you typically drive in the mud and need lots of car washes for a monthly fee.
\newline

Example: One hour wait for mediocre food. But at least the place pumps uber loud music so everyone had to scream to be heard.

Counterfactual: No wait for delicious food. The place plays music at the right volume so everyone could have to talk without any need to raise their voices.
\newline

Example: Excellent and fresh ingredients, make this a must go to for tasty sushi. Staff is unfriendly, but restaurant is spacious.

Counterfactual: Mediocre and stale ingredients, make this a place to avoid for tasty sushi. Although the staff is friendly, the restaurant is cramped.
\newline

Example: Nice place. Quick and easy. Had the eggs and corned beef hash special - which was great. Would come back to try more. Coffee was not good - especially with so many good coffee options in the Strip.

Counterfactual: Awful place. Slow and complicated. Had the eggs and corned beef hash not special - which was terrible. Would never come back to try more. Coffee was surprisingly good - especially with so many bad coffee options in the Strip.
\newline

Example: It's pretty much better than you expect for the money. Nothing to complain in terms of food and in comparison to barbarians it is more affordable... And they even have pickled vegetables for appetizers :)

Counterfactual: It's pretty much worse than you expect for the money. Plenty to complain about in terms of food and in comparison to barbarians it is much more expensive... And they don't even have pickled vegetables for appetizers :(
\end{framed}

\begin{table*}[h]
    \centering
    \begin{tabular}{p{5cm}p{5cm}p{5cm}}
    \toprule
    Source Text & Counterfactual(Basic Prompt) & Counterfactual(Optimized Prompt) \\
    \midrule
\textcolor{blue}{Great place to eat. Food always tastes fresh. Frequently visit ray road location. Ice machine always seems to be working. Very sanitary to scoop ice with a clean scooper provided.} & \textcolor{blue}{Great place to eat. Food always taste fresh. Frequently visit ray road location.} \textcolor{red}{Ice machine always seems to be broke. Kind of unsanitary to scoop ice with scooper laying in the ice.} & \textcolor{red}{Terrible place to eat. Food always tastes stale. Rarely visit ray road location. Ice machine never seems to be working. Very unsanitary to scoop ice with a dirty scooper provided.} \\
    \midrule
\textcolor{blue}{I've been here before...amazing food ...and the consistency is always on point...service was exceptional...love this place cause it is close to the house...Jose's bandana must be giving him superpowers when he's cooking!!} & \textcolor{blue}{I've been here before...good food} ...\textcolor{red}{but the consistency needs improvement...service was mediocre}...\textcolor{blue}{still like this place cause it is close to the house...} \textcolor{red}{maybe Jose's bandana is covering his eyes when he's cooking!!} & \textcolor{red}{I've been here before...terrible food ...and the inconsistency is always a problem...service was terrible...hate this place cause it is far from the house... Jose's bandana must be giving him bad luck when he's cooking!!} \\
    \midrule
\textcolor{blue}{It's great. Floor was clean and our waiter seemed knowledgeable. Food was excellent and definitely worth writing home about.} & \textcolor{blue}{It's ok.} \textcolor{red}{ Floor was dirty and our waiter seemed almost clueless.} \textcolor{blue}{Food was good but not "write home about"} & \textcolor{red}{It's terrible. The floor was dirty and our waiter seemed clueless. The food was awful and definitely not worth mentioning.} \\
    \bottomrule
    \end{tabular}
    \caption{Some of the counterfactuals generated under the Basic Prompt and Optimized Prompt settings. \textcolor{red}{red} part represents negative, and the \textcolor{blue}{blue} is positive.}
    \label{tab:examples}
\end{table*}
\end{document}